\pdfoutput=1

\documentclass[11pt]{article}

\usepackage{EMNLP2023}

\usepackage{times}
\usepackage{latexsym}
\usepackage{makecell}

\usepackage[T1]{fontenc}

\usepackage[utf8]{inputenc}

\usepackage{microtype}
\usepackage{adjustbox}
\usepackage{booktabs}
\usepackage{amsmath}
\usepackage{bbm}

\usepackage{inconsolata}

\usepackage{graphicx}
\usepackage{tipa}
\usepackage{array}

\usepackage{xcolor}
\usepackage{soul}
\newcommand{\hlc}[2][yellow]{{%
    \colorlet{foo}{#1}%
    \sethlcolor{foo}\hl{#2}}%
}
\definecolor{babyblue}{rgb}{0.54, 0.81, 0.94}
\definecolor{babypink}{rgb}{0.96, 0.76, 0.76}
\definecolor{brilliantlavender}{rgb}{0.96, 0.73, 1.0}

\newcommand{\namedref}[2]{\hyperref[#2]{#1~\ref*{#2}}}

\newcommand{\tableref}[1]{\namedref{Table}{#1}}
\newcommand{\figureref}[1]{\namedref{Figure}{#1}}
\newcommand{\appendixref}[1]{\namedref{Appendix}{#1}}
\newcommand{\knnlm}[0]{$k$NN-LM}
\newcommand{\knnlms}[0]{$k$NN-LMs}

\title{$k$NN-LM Does Not Improve Open-ended Text Generation}

\author{
  Shufan Wang$^1$\hspace{3mm} Yixiao Song$^1$\hspace{3mm} Andrew Drozdov$^1$ \hspace{3mm}\\ \textbf{Aparna Garimella}$^2$\hspace{3mm} \textbf{Varun Manjunatha}$^2$\hspace{3mm} \textbf{Mohit Iyyer}$^1$\vspace{0.2em}\\
  University of Massachusetts Amherst$^1$ \hspace{1em}Adobe Research$^2$ \hspace{1em} \\
  \texttt{\{shufanwang, yixiaosong, adrozdov, miyyer\}@umass.edu} \\
  \texttt{\{garimell,vmanjuna\}@adobe.com} \\
  } 

\begin{document}
\maketitle
\begin{abstract}





In this paper, we study the generation quality of interpolation-based retrieval-augmented language models (LMs). These methods, best exemplified by the \knnlm~\citep{khandelwal20generalization}, interpolate the LM's predicted distribution of the next word with a distribution formed from the most relevant retrievals for a given prefix. While the \knnlm\ and related methods yield impressive decreases in \emph{perplexity}, we discover that they do not exhibit corresponding improvements in \emph{open-ended generation quality}, as measured by both automatic evaluation metrics (e.g., MAUVE) and human evaluations. Digging deeper, we find that interpolating with a retrieval distribution actually \emph{increases} perplexity compared to a baseline Transformer LM for the majority of tokens in the WikiText-103 test set, even though the overall perplexity is lower due to a smaller number of tokens for which perplexity dramatically decreases after interpolation. However, when decoding a long sequence at inference time, significant improvements on this smaller subset of tokens are washed out by slightly worse predictions on most tokens. Furthermore, we discover that the entropy of the retrieval distribution increases faster than that of the base LM as the generated sequence becomes longer, which indicates that retrieval is less reliable when using model-generated text as queries (i.e., is subject to exposure bias). We hope that our analysis spurs future work on improved decoding algorithms and interpolation strategies for retrieval-augmented language models. 

\end{abstract}

\section{Introduction}














Retrieval-augmented language models, which integrate non-parametric dense retrieval with autoregressive next-token prediction, have been validated with strong empirical performance across a variety of tasks \cite{metzler2022reml,Basu2022GeneralizationPO,Mialon2023AugmentedLM} in addition to achieving low held-out perplexities on LM benchmarks. In this paper, we study \emph{interpolation-based} LMs, a subtype of retrieval-augmented LMs that compute the probability of the next token by interpolating between the softmax distribution of the original LM and a token distribution formed by retrieving over an external datastore. These methods, perhaps best exemplified by the \knnlm~\cite{khandelwal20generalization}, are particularly attractive because they allow any pretrained LM to be retrofitted with a retrieval module without further training.

Despite these advantages, there is limited understanding about the \emph{text generation quality} of interpolation-based LMs. In this study, we evaluate the quality of generated text from two such methods, \knnlm\ and TRIME~\cite{Zhong2022TrainingLM}, against the output of baseline LMs that do not use retrieval. Our evaluations involves \emph{open-ended} text completions generated using different decoding algorithms on the WikiText-103 dataset. We discover that interpolation-based LMs do not improve the quality of generated text, as measured by both automatic text generation metrics such as MAUVE~\citep{Pillutla2021MAUVEMT} and human evaluation. 

This result begs the question of \emph{why} the text generation quality does not improve, as the perplexity of interpolation-based LMs is substantially lower than that of the baselines.  Our analysis of the \knnlm\ model suggests two potential reasons for this lack of improvement:

\begin{enumerate}
    \item \knnlm\ actually \emph{worsens} the predictions of the majority of tokens in the WikiText-103 test set. On aggregate, perplexity improves because of significantly improved predictions on a smaller subset of tokens. However, when generating a long sequence of tokens, these improvements are washed out by the worsened predictions on other tokens. 
    \item The quality of the retrieval distribution deteriorates faster than that of the LM's predicted distribution as the length of the generation increases; in other words, the retrieval distribution is more vulnerable to exposure bias and can be easily thrown off by artifacts presented in model-generated text. 
\end{enumerate}

Unlike previous works that rely on perplexity to evaluate language modeling or BLEU to evaluate machine translation quality of \knnlm-based models \cite{khandelwal2021nearest}, our work specifically studies the open-ended text generation capability of \knnlms\ with a range of automatic evaluation metrics as well as human evaluation.
We demonstrate that, though they significantly lower perplexity, retrievers might also impair text generation performance of \knnlms. This finding suggests potential future directions for using retrieval during text generation, such as developing more robust retrieval components or employing retriever mechanisms more selectively during decoding.

\section{Related Work}

We present the most extensive study of open-ended text generation\footnote{The $k$NN-LM is also evaluated using MAUVE in \citet{lan2023copy}; however, our work has much more extensive analysis in the open-ended text generation setting.} from interpolation-based LMs such as $k$NN-LM \cite{khandelwal20generalization}.  Our results reveal that although these methods are effective at reducing perplexity, they can also be detrimental to text generation. Previous work finds that retrieval LMs are improved by selectively incorporating retrieval when conditions are favorable \cite{he-etal-2021-efficient,alon2022neuro,drozdov-etal-2022-cant,mallen2023when}, although they only examine the teacher-forced setting or other tasks, e.g. question answering. The $k$NN-MT \cite{khandelwal2021nearest} explores machine translation, which is a constrained task with short inputs, and thus not a good test of open-ended long-form generation.

The $k$NN-LM effectively scales retrieval to billions of tokens using a token-level non-parametric interpolation technique first introduced by \citet{grave2017improving}. Alternative retrieval-augmented models experiment with training the retriever \cite{Zhong2022TrainingLM,ram2023ralm,Shi2023REPLUGRB}, interpolating vectors instead of token probabilities \cite{Yogatama2021AdaptiveSL}, scaling to trillions of tokens \cite{Borgeaud2021ImprovingLM},  exploiting retrieval for strong few-shot learning \cite{izacard_few-shot_2022}, and so on \cite{chen-etal-2017-reading,guu2020realm,lewis2020rag,izacard-grave-2021-leveraging,Rae2021ScalingLM,wu2022memorizing,Trivedi2022InterleavingRW,He2022RethinkingWR}. Among these, \knnlm\ stands out as a relatively simple and fundamental work. Our findings indicate important weaknesses of retrieval for text generation.

Reference-based metrics are not well suited to evaluate open-ended text generation \cite{novikova-etal-2017-need}.
Instead, effective automated approaches compare the machine generated and human language text distributions using samples \cite{McCoy2021HowMD,Pillutla2021MAUVEMT,pimentel2023on}. 
Human evaluation remains the golden standard for natural language generation \cite{hashimoto-etal-2019-unifying,celikyilmaz2020evaluation,Krishna2023LongEvalGF}.

\section{Experimental setup}




















Using a variety of commonly used text generation evaluation metrics, we evaluate the text generation capability of interpolation-based LMs and compare them to baseline LMs (i.e., without $k$-nearest-neighbor retrieval from an external datastore). In this section, we describe our experimental setup, including models, automatic evaluation metrics, data selection, and hyperparameters.

\subsection{Models}
We experiment with two interpolation-based LMs: the \knnlm\ of~\citet{khandelwal20generalization}, which augments an existing pretrained LM with a retrieval module without any additional training, and TRIME~\cite{Zhong2022TrainingLM}, a recent improvement over the \knnlm\ that trains the retriever and LM jointly to further decrease perplexity. 

\paragraph{\knnlm:} The \knnlm\ is a pretrained language model that uses retrieval to improve word prediction. We follow the procedure from~\citet{khandelwal20generalization}\footnote{Alternative distance functions, token representations, and interpolation options for \knnlm\ are explored in \citet{Xu2023WhyDN}. 
We don't expect those settings to impact the trends we observe, but as we mention in \S\ref{sec:discussion}, tuning for text generation could be beneficial.} and use the LM to encode token-level representations from a document collection (e.g., WikiText-103 training data) into a datastore where each token in document is converted into a key-value pair: a context vector $k_i$ representing the first $n-1$ words and a value $v_i$ which is the $n$-th word. During evaluation, the model calculates Euclidean distances $d(k, q_j)$ between the query vector $q_j$ and all the keys $k_1, k_2, \dots k_{|V|}$ in the datastore. The values from the retrieved documents define a new distribution of the next word:

\begin{equation}
    P_{KNN}(w_t | q_t) \propto \sum_{(k_i, v_i)} \mathbbm{1}_{w_t = v_i} \exp(-d(k_i, q_t))
\end{equation}

\noindent The model interpolates the LM's predicted distribution over the next token $P(w_t | q_t)$ with the retrieval distribution with a tunable hyperparameter $\lambda$:

\begin{equation}\label{eq:knnlm}
    P^{'}(w_t | q_t) = \lambda P_{KNN}(w_t | q_t) + (1-\lambda) P_{LM}(w_t | q_t)
\end{equation}

To generate text from the \knnlm, we apply a decoding strategy (e.g., greedy decoding or truncated sampling algorithms) using the final interpolated probability distribution  $P^{'}(w_t | q_t)$.

\paragraph{TRIME:} Note that in \knnlm, the LM is trained \emph{without} retrieval; the retrieval component is bolted on after training.~\citet{Zhong2022TrainingLM} note that this approach is suboptimal, as the LM does not understand how to best use the retrieval. Thus, they propose the TRIME model, which uses an efficient in-batch strategy to incorporate retrievals during training. While \knnlm\ relies on just one type of retrieval (from an external datastore), TRIME can retrieve from local and long-range context as well as external context. We use the TRIME$_{\text{EXT}}$ configuration in all of our experiments, which also uses a linear interpolation between LM and retrieval distributions (as in Equation~\ref{eq:knnlm}) to produce the final probability distribution. The baseline LM (no external retrieval) can still retrieve from example-level local and long context, but it has no access to a huge-scale external datastore.



\subsection{Constructing an evaluation dataset}
We sample from WikiText-103~\citep{merity2016pointer} to construct an evaluation dataset. We choose WikiText-103 because it is the most commonly used dataset for evaluating interpolation-based LMs; indeed, the main experiments from both \knnlm\ and TRIME demonstrate that the retrieval component decreases held-out perplexity on this dataset compared to the baseline LM. Specifically, we randomly sample 5K examples\footnote{We choose 5K examples because this is the minimum recommended number of generations to obtain meaningful comparisons as per ~\citet{Pillutla2021MAUVEMT}.} from the validation and test set of WikiText-103, and  we use the first 100 tokens of each example as a \emph{prefix} that the model must condition on to generate a 150-token-long continuation. As some of our metrics requires reference text, we also store the ground-truth 150 tokens (\emph{gold suffix}) that follow the prefix in each example.

\subsection{Automatic evaluation metrics}
For both \knnlm\ and TRIME, we compare the quality of text generated by the base LM with and without the $k$-NN retrieval component over the external datastore. We measure quality via the following automatic metrics:

\paragraph{MAUVE:} MAUVE is an evaluation metric for open-ended text generation \citep{Pillutla2021MAUVEMT} that achieves high correlation with human judgments of text quality. It measures the distribution similarity between the generated text and the reference text. Higher MAUVE scores indicate closer distance between the distribution of the generated text and that of reference text.

\paragraph{RankGen:} Given a prefix and several possible continuations (suffixes), RankGen~\citep{krishna-etal-2022-rankgen} outputs a score for each suffix, measuring the relevance between the prefix and suffix. Higher RankGen scores indicate stronger relevance between generated suffix with the given prefix. We thus measure the RankGen score between prefix and generated suffix for each of the two models. 

\paragraph{GPT-3 perplexity:} We also use GPT-3~\citep{NEURIPS2020_1457c0d6}, a large-scale pretrained language model, to compute the perplexity of text generated with and without interpolation conditioned on the same prefix. Lower GPT-3 perplexity indicates stronger relevance between the prefix and generated suffix and the better fluency of the generated suffix. We use the 6.7B \texttt{gpt3-curie} model via OpenAI's API to measure perplexity.

\paragraph{Entity-F1:} Previous works \citep{nan-etal-2021-entity, lee2022factuality} use the percentage of hallucinated named entities (entities that appear in the generated text but not in the reference text) or the ratio of named entity overlaps between the generated text and reference text to estimate the factuality of the generated text. In our work, we compute the F1 scores between the named entities from the generated text and reference text as a proxy for entity hallucination. Higher F1 scores may correlate to fewer instances of hallucinated entities.

\paragraph{Seq-Rep-1:} We follow \citet{unlikelihood-welleck} and use the percentage of unique unigrams (Seq-Rep-1) in the text as a metric for lexical diversity in the text. Higher Seq-Rep-1 scores indicate lower diversity (more repetition) in the generated text.

\subsection{Model configurations and hyperparameters}
In this work, we do not train our own interpolation-based LMs but rather leverage pretrained model and datastore checkpoints released by prior work.

\paragraph{Base LM details:}
For \knnlm, we use the implementation from~\citet{alon2022neuro}, which relies on a backbone 117M-parameter GPT-2 small model~\citep{radford2019language} fine-tuned on the WikiText-103 training data. The external datastore is constructed by the same backbone model, and both the pretrained LM and datastore are publicly released by~\citet{alon2022neuro}.\footnote{See the \texttt{gpt2-finetuned-wikitext103} model available here: \url{https://github.com/neulab/knn-transformers}.} 
For TRIME, we use the 247M-parameter TRIME$_{\text{ext}}$ model trained from scratch on WikiText-103 and publicly released by~\citet{Zhong2022TrainingLM}. Our ``non-retrieval'' baseline is the same model without external retrieval; in other words, it has  access to only the local memory (recent tokens) and long-range memory (in-batch tokens). In both the \knnlm\ and TRIME setups, the external datastore is constructed using the training dataset of WikiText-103; the TRIME datastore size is 103M entries, while the \knnlm\ has 117M entries (the discrepancy is due to tokenization differences between the two models). 

\paragraph{Perplexity improvements from retrieval:} Both models studied in this paper substantially decrease perplexity on WikiText-103's validation set when interpolation is enabled. For \knnlm, the base GPT-2 perplexity is 14.8, and it decreases to 12.6 (-2.2) after interpolation. Meanwhile, TRIME decreases perplexity from 17.0 (no retrieval) to 15.5 (-1.5) after interpolation.

\paragraph{Hyperparameters:}
To generate text, we use the hyperparameters recommended by the authors that yield low perplexities on the WikiText-103 test set. For the \knnlm, the softmax temperature is set to $1.0$ and the interpolation coefficient between the LM distribution and the retrieval distribution $\lambda$ is set to $0.25$. For TRIME, the softmax temperature is set to $1.25$ and the $\lambda$ is $0.3$. For most of our experiments (e.g., those in Table~\ref{tab:eval_metrics}), unless otherwise specified, we decode the continuations using nucleus sampling~\citep{holtzman2019curious} with $p=0.8$.

\section{Results}

We find that despite incorporating the retrieval component and interpolating the information from the base-LM and the retrieval, these methods do not yield any significant improvement to text generation performance, and  even worsen it by some metrics (Table \ref{tab:eval_metrics}). In this section, we provide an overview of our main results, perform more fine-grained analyses, and describe a human evaluation that supports the conclusions drawn from automatic metrics.

\begin{table}[t]
    \centering
    \small
    
    \begin{adjustbox}{max width=0.50\textwidth}

        \begin{tabular}{p{1.9cm}p{0.7cm}p{0.7cm}p{0.7cm}p{0.7cm}p{0.7cm}}
        \toprule
          \bf Model & \bf \tiny MAUVE$\uparrow$ & \bf \tiny PPL$_\text{GPT-3}$$\downarrow$ &\bf \tiny RankGen$\uparrow$  & \bf \tiny EntityF1$\uparrow$ & \bf \tiny SeqRep\_1$\downarrow$ \\
          \midrule
            \multicolumn{6}{l}{\scriptsize{\emph{\knnlm\ with and without retrieval from~\citet{alon2022neuro}}}}\vspace{0.1cm} \\

           \begin{tabular}{@{}l@{}}GPT-2 small\\[-0.3ex]{\footnotesize \textit{(no retrieval)}}\end{tabular} & 0.773 
            & 13.1
            & 11.7
            & 0.14
            & 0.57 \vspace{0.3cm} \\
            
           \begin{tabular}{@{}l@{}}GPT-2 small\\[-0.3ex]{\footnotesize \emph{(+ retrieval)}}\end{tabular} & 0.793
           & 14.8
            & 11.7
            & 0.13
            & 0.53 \\\midrule
            \multicolumn{6}{l}{\scriptsize{\emph{TRIME$_\text{EXT}$ with and without external retrieval from~\citet{Zhong2022TrainingLM}}}}\vspace{0.1cm} \\
            
          \begin{tabular}{@{}l@{}}TRIME\\[-0.3ex]{\footnotesize \emph{(no ext retrieval)}}\end{tabular} & 0.889
            & 23.1
            & 13.0
            & 0.09
            & 0.40 \vspace{0.3cm} \\
            
          \begin{tabular}{@{}l@{}}TRIME\\[-0.3ex]{\footnotesize \emph{(+ ext retrieval)}}\end{tabular}
            & 0.885
            & 24.7
            & 12.3
            & 0.08
            & 0.39 \\

          \bottomrule
        \end{tabular}
        \end{adjustbox}
    \caption{Automatic evaluation metrics do not show consistent improvement in generation quality for interpolation-based LMs---\knnlm\ (top), and TRIME (bottom)--- compared to no-retrieval baseline LMs.}
    \label{tab:eval_metrics}
\end{table}

\paragraph{Interpolation-based LMs do not improve automatic text generation evaluation metrics:} We find that neither \knnlm\ nor TRIME significantly improve generation quality compared to the base LM, as shown by various evaluation metrics (Table \ref{tab:eval_metrics}). For \knnlm, while the MAUVE score improves by 2 points with retrieval, the perplexity of GPT-3 \emph{increases} on retrieval-augmented generations, and the RankGen score is identical. For TRIME, the no-retrieval baseline is actually slightly \emph{better} across MAUVE, GPT-3 perplexity, and RankGen. In other words, there is no convincing winner; furthermore, 
contrary to the expectation that \knnlms\ may reduce hallucination by retrieving (and potentially copying) from the datastore, we also do not observe any improvement in the Entity F1 scores with the gold suffix. We observe a marginal (likely insignificant) improvement in lexical diversity of the generations (shown by the lower seq\_rep\_1 score). 

\paragraph{These results hold across different decoding algorithms:}
The results in Table~\ref{tab:eval_metrics} are all from nucleus sampling. What if we change the decoding algorithm? To investigate the impact of decoding algorithm on generation quality, we evaluate the \knnlm\ on three different decoding algorithms: greedy decoding, ancestral sampling, and beam search. 
We observe in Table \ref{tab:decoding} that none of these decoding algorithms changes the result: there is no clear winner between models with and without retrieval.

\begin{table}[t]
    \centering
    \small
    
    \begin{adjustbox}{max width=0.50\textwidth}

        \begin{tabular}{p{1.8cm}p{1.5cm}p{1.5cm}p{1.5cm}}
        \toprule
          \bf Model 
          & \bf \small Nucleus Sampling 
          & \bf \small Top-$k$ \newline Sampling 
          & \bf \small Beam Search
          \\
          \midrule
            \multicolumn{4}{l}{\scriptsize{\emph{\knnlm\ with and without retrieval from~\citet{alon2022neuro}}}}\vspace{0.1cm} \\

           \begin{tabular}{@{}l@{}}GPT-2 small\\[-0.3ex]{\footnotesize \textit{(no retrieval)}}\end{tabular} & 0.773 
            & 0.807
            & 0.0363 \vspace{0.3cm} \\
            
           \begin{tabular}{@{}l@{}}GPT-2 small\\[-0.3ex]{\footnotesize \emph{(+ retrieval)}}\end{tabular} & 0.793
           & 0.793
            & 0.0338 \\ \bottomrule
        \end{tabular}
        \end{adjustbox}
    \caption{The observation that \knnlm\ does not significantly improve text generation performance (measured here via MAUVE) is consistent across a variety of decoding algorithms: nucleus sampling, top-$k$ sampling ($k=40$) and beam search (beam size $=5$). We note that beam search decoding often generates repetitive text and therefore scores poorly with MAUVE.}
    \label{tab:decoding}
\end{table}

\subsection{Human evaluation}


Having found that interpolation-based LMs do not notably improve text generation quality according to automatic evaluation metrics, we turn next to human evaluation, which is known to be more reliable for generation tasks~\citep{celikyilmaz2020evaluation,krishna-etal-2021-hurdles}, to compare the text generated by the \knnlm\ vs. the baseline GPT-2 model. We hired three English teachers/editors on the freelance marketplace Upwork. The evaluation was conducted on the platform Label Studio~\citep{LabelStudio}.\footnote{\url{https://www.upwork.com}, \url{https://labelstud.io/}} The annotators were experienced in text generation evaluation and hired after careful selection.

The annotators were given a prefix and two continuations of the context (one generated by the baseline LM and one generated with retrieval). The presentation order of the two continuations were randomized. The evaluators' task was to decide which continuation is better, indicate whether it was hard to choose between the two following~\citet{thai-etal-2022-exploring}, and justify their choice in 3 to 4 sentences.\footnote{A screenshot of our evaluation platform can be found in \appendixref{appendix:humen_eval}.} The evaluation focused on whether the generated text is grammatical, fluent, consistent, and logical. Each evaluator evaluated 45 pairs of continuations generated by \knnlm~and GPT-2. 
Each evaluator was paid \$50 for their work.

\begin{figure}[t!]
    \centering
    \includegraphics[scale=0.39]{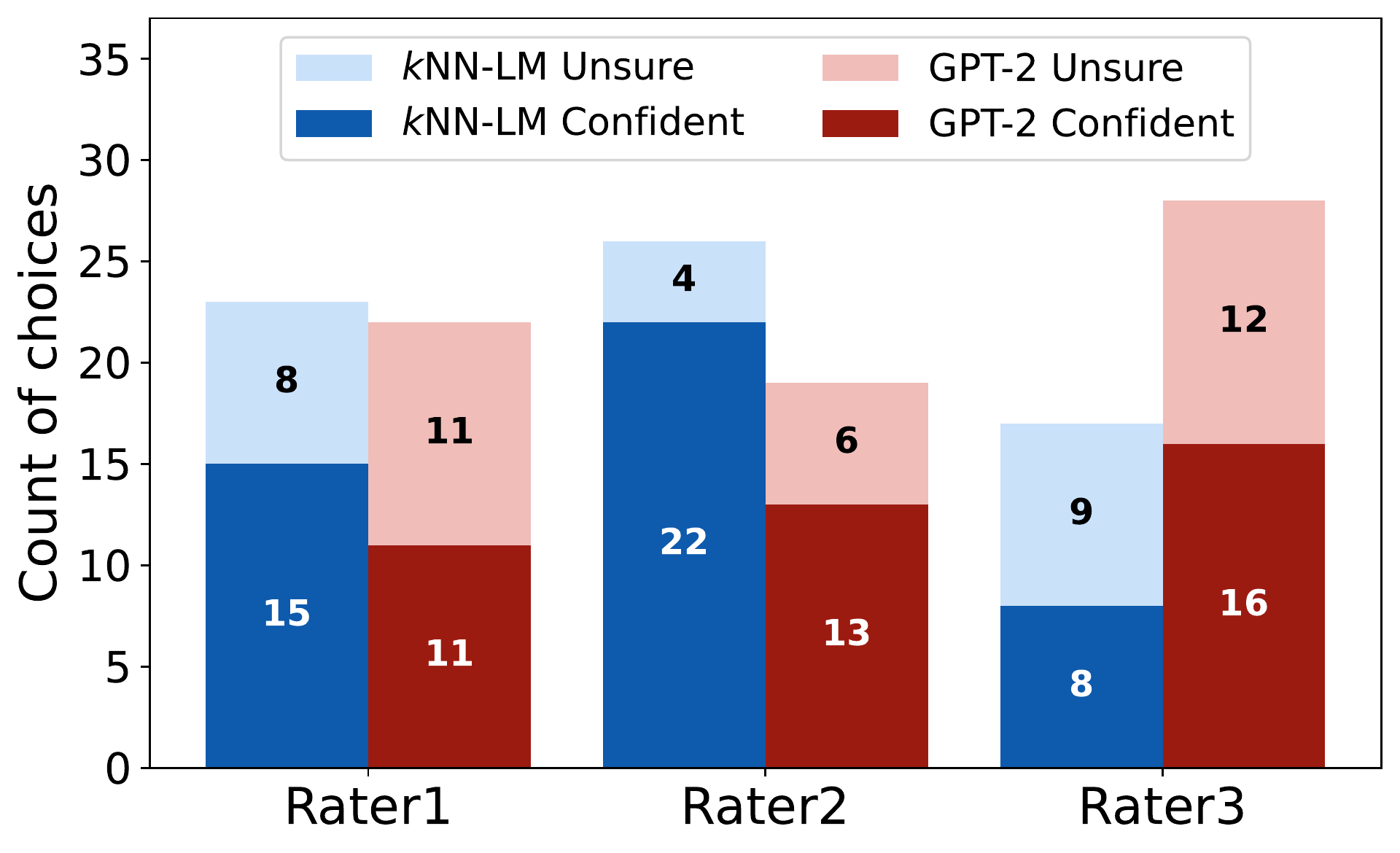}
    \caption{The plot presents how many times each type of generations (\knnlm~or GPT-2) is chosen by the evaluators. The dark area in each bar shows that the choices were made confidently. The light area represents the choices between \knnlm~and GPT-2 that were hard but the evaluator still chose the corresponding type. Overall, annotators preferred GPT-2 baseline texts $51\%$ of the time compared to 49\% for \knnlm.}
    \label{fig:human_eval_plot}
\end{figure}

\paragraph{Human evaluation shows no definitive winner between \knnlm~and GPT-2 either:} On aggregate, baseline GPT-2 generations were preferred $51\%$ of the time, vs. $49\%$ for \knnlm. Additionally, the three annotators report that the decision was difficult for $37\%$ of all cases. Out of the 45 comparison pairs, the three annotators only agree on their choices in 17 instances ($37.78\%$), resulting in a Fleiss Kappa score $0.17$ (slight agreement). \figureref{fig:human_eval_plot} presents the evaluator preference when comparing the \knnlm~to GPT-2 generations. The light area shows the choices that were hard to make but the evaluator still chose the corresponding type. For Rater1 and Rater3, the rates of \textit{difficult to choose} are as high as $42\%$ and $47\%$ while for Rater2 it is $22\%$. 

\paragraph{Both models make catastrophic errors at similar rates:}
A qualitative analysis of the free-form choice justifications from the evaluators reveals that both \knnlm~and GPT-2 make catastrophic mistakes. \tableref{tab:human_eval_error} gives four examples of bad continuations, along with the evaluators' comments and our categorization of the errors. In the first row of the table, Continuation A generated by the \knnlm~contains repetitive content (i.e., \textit{==ZAPU retreat==}), and confuses \textit{ZAPA} and \textit{ZIPRA} at multiple places. The GPT-2 continuation in the second row states that a person was born in 1584 but was still alive in 1742; the generation in the third row by the \knnlm~claims that U.S.\ Route 75 curves both northeast and northwest in the northbound direction. Furthermore, both the GPT-2 and \knnlm's generations change topics abruptly as shown in the lower half of \tableref{tab:human_eval_error}. 
Overall, the quantitative and qualitative analyses of the human evaluation results show that the \knnlm~does not clearly improve over its base GPT-2 model despite its significant improvement in perplexity.


\section{Why do \knnlms\ fail to improve text generation quality?}
\label{sec:why}

Our evaluations (both human and automatic) do not show a significant quality increase when interpolating an LM's predicted probability distribution with one formed via retrieval over a large external datastore. In this section, we try to understand \emph{why} we do not observe an improvement by empirically analyzing the \knnlm. We come up with two reasons: (1) despite lowering the aggregate perplexity, $k$NN-LMs only improve the perplexity of 42\% of all test tokens, which suggests that the improved quality of a subset of tokens could be counter-balanced by worsened predictions on other tokens that do not benefit from the \knnlm. 
Moreover, we find the entropy of the retrieval distribution to increase at a faster rate compared to that of the baseline LM as the model generates longer sequences. This difference implies that the retriever distribution is getting noisier as more tokens are sampled, potentially due to the exposure bias stemming from the retriever having to rely on the sampled text as the query.

\begin{figure}[t!]
    \centering
    \includegraphics[scale=0.5]{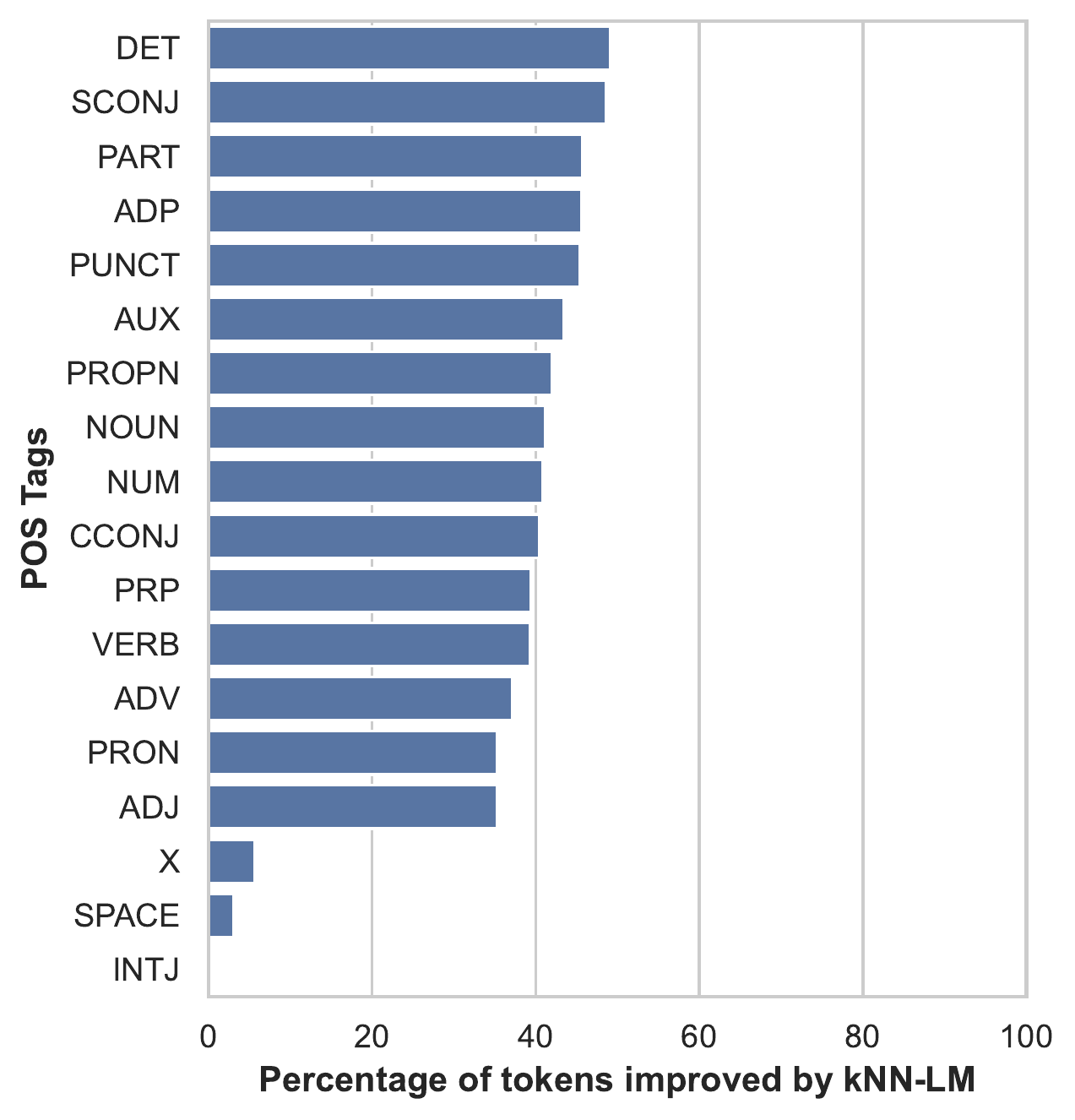}
    \caption{Across all POS tags, we observe that \knnlm\ does not increase the probability of the majority of gold next token predictions. For verbs, pronouns, and adjectives, it only helps $<40\%$ of the time (i.e., it hurts the predictions of the majority of these tokens).}
    \label{fig:pos_plot}
\end{figure}

\subsection{KNN-LMs only benefits a subset of tokens} 
Many studies have shown that \knnlms\ decrease perplexity via retrieval interpolation \citep{khandelwal20generalization, alon2022neuro, drozdov-etal-2022-cant}. Previous work~\citep{drozdov-etal-2022-cant, Zhong2022TrainingLM} has also suggested that \knnlms\ benefit the inference of tokens of various part-of-speech (POS) tags to different degrees (by lowering the perplexity of the gold token). However, these works focus on \textbf{aggregate} perplexity averaged across tokens in the testing examples but do not look at \textbf{individual} tokens and the percentage of tokens that actually benefit from retrieval. 

Using the dataset we selected from WikiText-103 for evaluating text generation, we compute the percentage of gold tokens from our test examples that are assigned lower perplexity (higher probability) by the \knnlm\ compared to the base LM. We find that only 42\% of the tokens benefit from \knnlms, while the remaining 58\% of the tokens are adversely affected by the \knnlm\ (i.e., the \knnlm\ assigns a smaller probability to the gold token compared to the baseline LM). Moreover, we also calculate the percentage of gold tokens that benefit from \knnlm\ in each POS category (Figure~\ref{fig:pos_plot}) and consistently find the similar result that \knnlm\ only helps reduce the perplexity for a smaller subset of tokens. We show examples of \knnlm\ negatively impacting the next-token prediction (assigning the gold token with lower probability compared to the base-LM) in Table \ref{tab:retrieval_examples}.

This means that despite lowering the \textbf{aggregate} perplexity across the test sets, the \knnlm\ is more likely to hurt, instead of help, the inference of each \textbf{individual} token. 
Therefore, we hypothesize that during text generation, as the model samples a sequence of tokens, the advantages brought by \knnlm\ to a smaller subset of tokens are offset by other tokens, for which \knnlm\ may even have a detrimental impact on the inference.

\subsection{The retriever becomes less reliable with longer generated sequences}\label{sec:exposure}
Additionally, we observe that as the model generates longer sequences of text, the retriever component from \knnlm\ becomes less confident and reliable in returning a high-quality next-token distribution. Since the \knnlm\ relies on interpolating the next-token distribution from the baseline LM and that from the retriever, a lower quality retriever distribution can compromise the resulting next-token distribution and adversely affect the text generation performance.

\begin{figure}[t]
    \centering
    \includegraphics[scale=0.45]{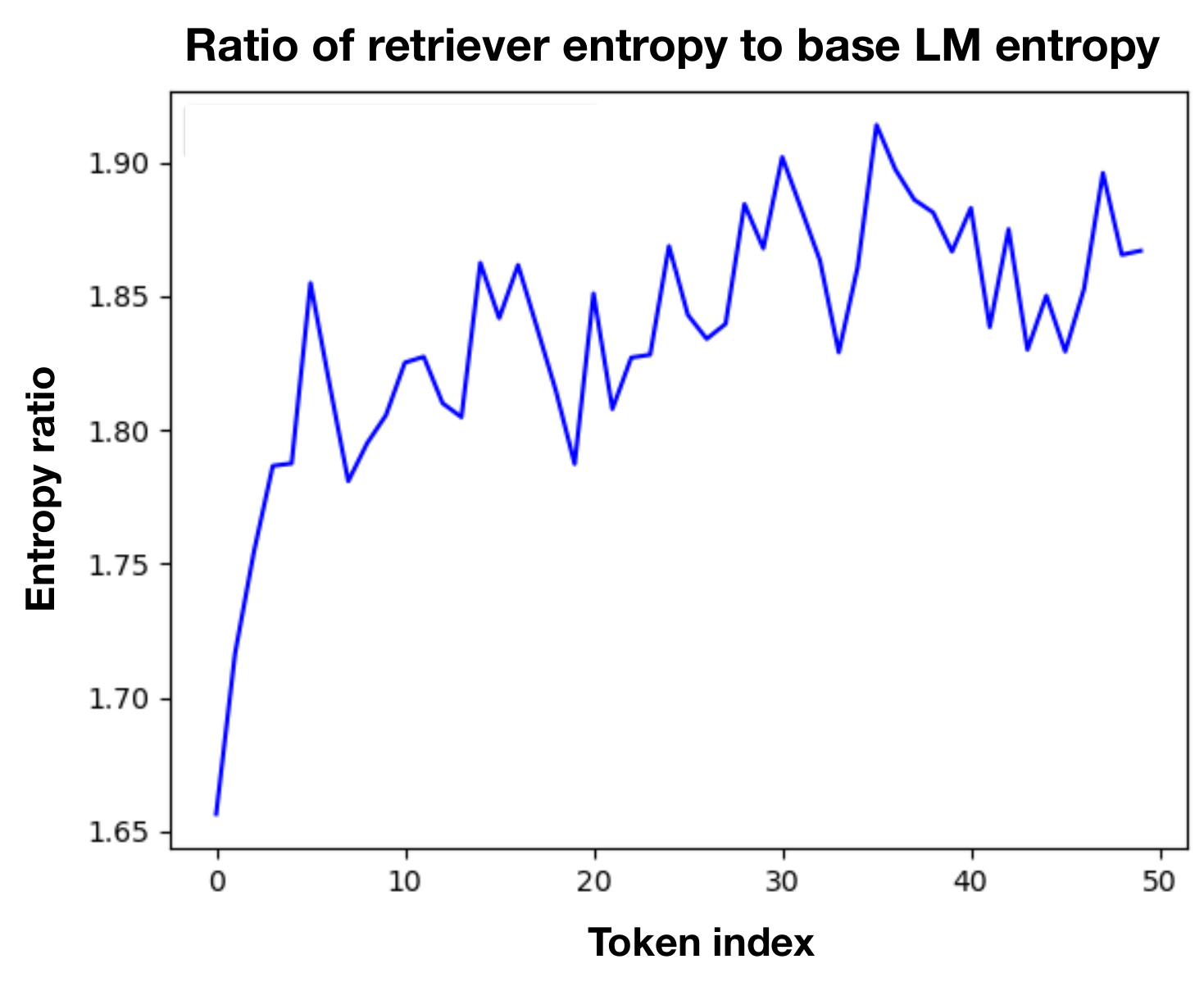}
    \caption{We plot the ratio between the Shannon entropy of the retriever's next-token distribution and that of the baseline LM softmax distribution, as the number of generated tokens increases. The ratio increases for longer model-generated sequences, indicating that the retriever becomes less confident than the baseline LM as decoding progresses.}
    \label{fig:self-entropy}
\end{figure}

\begin{figure}[t]
    \centering
    \includegraphics[scale=0.45]{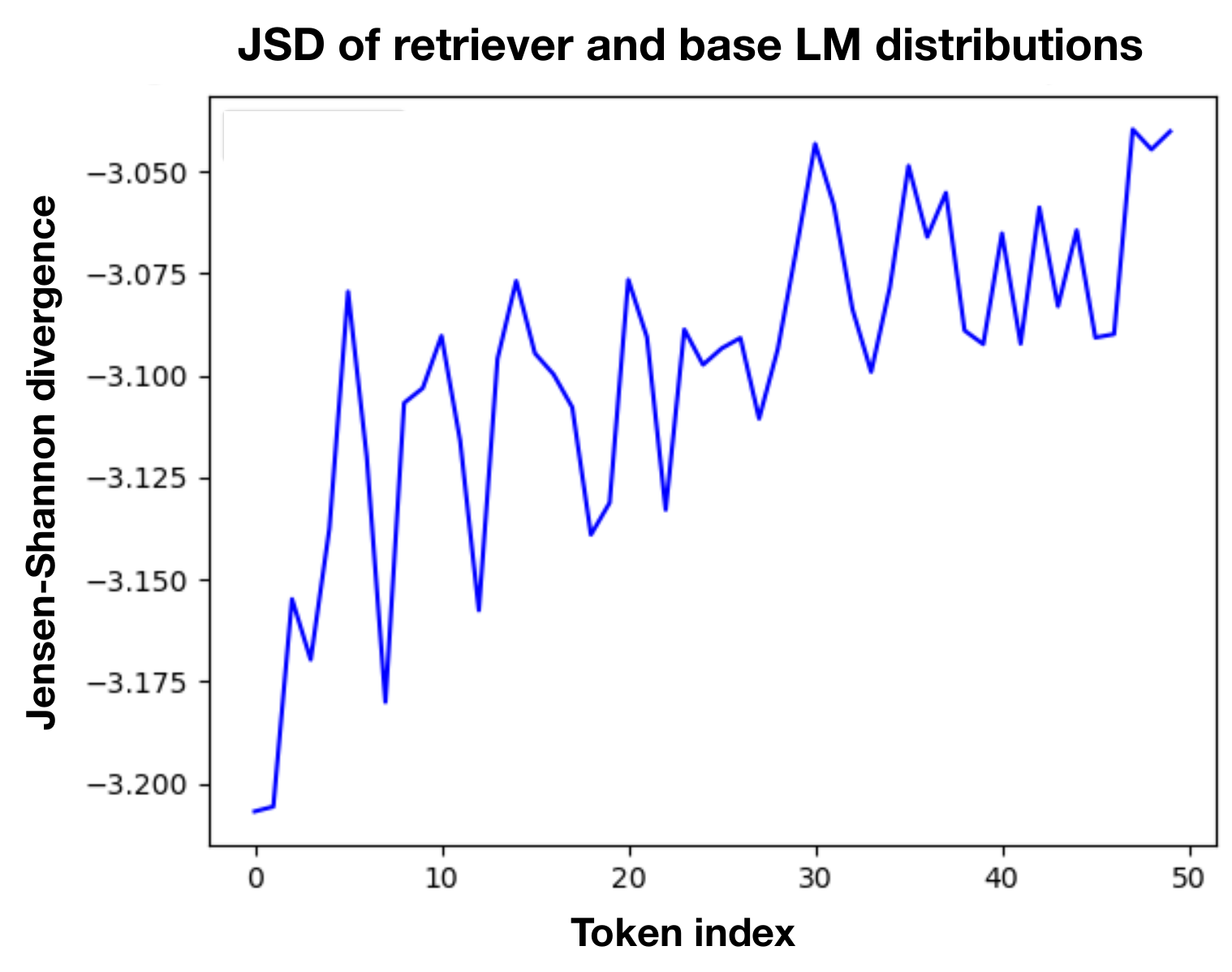}
    \caption{We plot the Jensen-Shannon divergence between the retriever's next-token distribution and that of the baseline LM softmax distribution, as the number of generated tokens increases. The increasing divergence indicates more disagreement between the retriever and the baseline LM in selecting the next token to generate.}
    \label{fig:js_div}
\end{figure}

We plot the ratio of Shannon entropy~\citep{shannon2001mathematical} between the retriever distribution and that of the baseline LM distribution on the next token (with respect to the index of the token generated) and find that the retriever's entropy is increasing at a faster rate compared to that from the base-LM (Figure~\ref{fig:self-entropy}). 
Given a $|V|$-dimensional probability distribution $p$, the entropy is computed as:
$$
H(p) = - \sum_{i=1}^{d} p_i \log(p_i)
$$
A higher entropy indicates lower level of confidence (closer to a uniform distribution over all tokens) and suggests that the retriever, when sampling long sequences, may be less reliable in identifying the high-quality tokens to be retrieved. 

Furthermore, we also plot the Jensen-Shannon probability distribution divergence between the retriever distribution and the baseline LM distribution over the next token, with respect to token indices. Given the retriever distribution $p$ and the baseline LM distribution $q$ (both $|V|$-dimensional), we calculate the Jensen-Shannon divergence ($D_{JS}$) as,

$$
D_{JS}(p | q) = \frac{1}{2} ( D_{KL}(p | m) + D_{KL}(q | m) )
$$
where $m$ is the mean distribution $\frac{1}{2} (p + q)$ and $D_{KL}(m)$ denotes the Kullback-Leibler divergence computed as $\sum_{i=1}^{d} p_i \log(\frac{p_i}{q_i})$

We observe that the probability distribution divergence between the retriever distribution and the base-LM distribution over the next-token widens as the sampled sequence becomes longer (~\ref{fig:js_div}), which means that they exhibit increased disagreement as more tokens are generated.

We hypothesize that the worsened reliability of the retriever over longer sampled sequences is likely a result of the \emph{exposure bias} during text generation (i.e., at test-time, the retriever has to rely on model-generated queries that may contain artifacts or other distributional differences from human-written text). The retriever in \knnlm\ is non-parametric since both the input prefix and the context from the datastore are encoded by the baseline LM (without any additional retrieval parameters), which has been adapted to the training corpus of WikiText-103. However, during text generation, as the model iteratively sample more tokens and append them to the input prefix, the input context is more likely to deviate from the available contexts from the training corpus and hence becomes more out-of-distribution and challenging for the retriever to accurately process.

\section{Discussion \label{sec:discussion}}
In addition to the limitations of interpolation-based LMs described in Section \ref{sec:why}, we hypothesize that there are other potential  factors that contribute to the shortcomings of \knnlm\ and TRIME for text generation. Specifically, it is possible that the interpolation may impede the language models' ability for self-recovery, and also that integrating the retrieval distribution can potentially introduce additional burdens related to hyperparameter tuning, which may not be optimized for text generation. We discuss these potential issues here as they are interesting avenues to explore for future work.

\paragraph{Retrieval interpolation may damage the self-recovery ability of LMs:} 
Language models exhibit some degree of self-recovery abilities \citep{he-etal-2021-exposure}, i.e., they can regain fluency and coherence even after previously generating poor-quality tokens. This self-recovery capability is attributed to the LM's ability to pay close attention to recent context and ignore information from the long-range history of past context.
However, we hypothesize that when interpolation-based LMs encounter artifacts (e.g., non-factual or disfluent text) in a distorted prefix $\Tilde{q}_t$, they may be less likely to recover than the baseline LMs, as the retrievals may further increase the probability of completions that resemble those artifacts. 
Furthermore, as we continuously sample tokens and append them to the prefix, which the retriever uses as the query to construct $P_{KNN}(w_t | \Tilde{q}_t)$, the retriever may encounter additional exposure bias as shown in Section~\ref{sec:exposure}, negatively impacting the quality of $P_{KNN}(w_t | \Tilde{q}_t)$.
Consequently, even when the baseline LMs ``recover'' from distorted past context by producing a high-quality distribution over the next-token prediction $P_{LM}(w_t | \Tilde{q}_t)$, the retriever may re-introduce the distortion by interpolating $P_{LM}(w_t | \Tilde{q}_t)$ with $P_{KNN}(w_t | \Tilde{q}_t)$.

\paragraph{Hyperparameters introduced by \knnlm\ are not optimized for text generation:} The \knnlm\ introduces two important hyperparameters, namely the relative weight between the two distribution $\lambda$, as well as softmax temperature for the $k$NN distribution $\tau_{KNN}$.  
Recent work \citep{Xu2023WhyDN}  highlights the significance of tuning $\tau_{KNN}$ for achieving optimal \knnlm\ performance, as measured by perplexity. 
Similarly, we hypothesize that the parameter $\lambda$ plays a vital role as it controls the relative importance assigned to the $k$NN retriever and the baseline LM, and instead of tuning $\lambda$ for optimizing perplexity, we may want to consider context-dependent $\lambda$ as in~\citet{drozdov-etal-2022-cant} for generation (e.g., only use the retrieval distribution when it is very confident). Finally, the interpolation may warrant the design of new decoding algorithms that are specialized for retrieval-augmented generation.

\section{Conclusion}

In this work, we show that despite the significant perplexity improvement brought by interpolation-based retrieval-augmented LMs such as \knnlms, such methods fail to improve the LMs' text generation performance. The text generation quality between \knnlms\ and baseline LMs without retrieval show no significant difference according to both automatic text generation evaluation metrics and human evaluation. Upon closer analysis, we identify flaws in using \knnlms\ to perform autoregressive text generation: the method only benefits a minority of token predictions, and the retriever's quality deteriorates when generating long-form text. We hope our findings can inspire future research to design better training and inference methods so that the impressive improvement of kNN-LMs in perplexity can better be translated into gains in text generation quality.


\section*{Limitations}
Our work does not study all data, model, and evaluation configurations of interpolation-based LMs.
We focus on Wikipedia text because it is the primary evaluation corpus for both \knnlm\ and TRIME. That said, it is unclear if our findings would be similar in other domains such as narrative or dialogue text, or in other languages. Additionally, we focus on the 100M token datastore size, although kNN-LM can scale effectively to datastores of 3B words. Using a larger datastore may lead to further perplexity decreases, but we do not think this contradicts our finding that text generation degrades as retrieval quality does.  We focus exclusively on interpolation-based LMs in this work, but similar issues for other retrieval-augmented LMs such as RETRO~\citep{Borgeaud2021ImprovingLM} may also exist and be worth investigating further.  Finally, our human evaluation does not specifically account for diversity, although some dimensions of this are captured by our automated metrics. Due to the overall low quality of text generated by LMs with and without retrieval, reading their outputs results in high cognitive burden on annotators, which might be ameliorated by using stronger LMs than GPT-2.

\bibliography{anthology,custom}

\begin{thebibliography}{46}
\expandafter\ifx\csname natexlab\endcsname\relax\def\natexlab#1{#1}\fi

\bibitem[{Alon et~al.(2022)Alon, Xu, He, Sengupta, Roth, and
  Neubig}]{alon2022neuro}
Uri Alon, Frank Xu, Junxian He, Sudipta Sengupta, Dan Roth, and Graham Neubig.
  2022.
\newblock Neuro-symbolic language modeling with automaton-augmented retrieval.
\newblock In \emph{International Conference on Machine Learning}, pages
  468--485. PMLR.

\bibitem[{Basu et~al.(2022)Basu, Rawat, and Zaheer}]{Basu2022GeneralizationPO}
Soumya~Sankar Basu, Ankit~Singh Rawat, and Manzil Zaheer. 2022.
\newblock Generalization properties of retrieval-based models.
\newblock \emph{ArXiv}, abs/2210.02617.

\bibitem[{Borgeaud et~al.(2021)Borgeaud, Mensch, Hoffmann, Cai, Rutherford,
  Millican, van~den Driessche, Lespiau, Damoc, Clark, de~Las~Casas, Guy,
  Menick, Ring, Hennigan, Huang, Maggiore, Jones, Cassirer, Brock, Paganini,
  Irving, Vinyals, Osindero, Simonyan, Rae, Elsen, and
  Sifre}]{Borgeaud2021ImprovingLM}
Sebastian Borgeaud, Arthur Mensch, Jordan Hoffmann, Trevor Cai, Eliza
  Rutherford, Katie Millican, George van~den Driessche, Jean-Baptiste Lespiau,
  Bogdan Damoc, Aidan Clark, Diego de~Las~Casas, Aurelia Guy, Jacob Menick,
  Roman Ring, T.~W. Hennigan, Saffron Huang, Lorenzo Maggiore, Chris Jones,
  Albin Cassirer, Andy Brock, Michela Paganini, Geoffrey Irving, Oriol Vinyals,
  Simon Osindero, Karen Simonyan, Jack~W. Rae, Erich Elsen, and L.~Sifre. 2021.
\newblock Improving language models by retrieving from trillions of tokens.
\newblock In \emph{International Conference on Machine Learning}.

\bibitem[{Brown et~al.(2020)Brown, Mann, Ryder, Subbiah, Kaplan, Dhariwal,
  Neelakantan, Shyam, Sastry, Askell, Agarwal, Herbert-Voss, Krueger, Henighan,
  Child, Ramesh, Ziegler, Wu, Winter, Hesse, Chen, Sigler, Litwin, Gray, Chess,
  Clark, Berner, McCandlish, Radford, Sutskever, and
  Amodei}]{NEURIPS2020_1457c0d6}
Tom Brown, Benjamin Mann, Nick Ryder, Melanie Subbiah, Jared~D Kaplan, Prafulla
  Dhariwal, Arvind Neelakantan, Pranav Shyam, Girish Sastry, Amanda Askell,
  Sandhini Agarwal, Ariel Herbert-Voss, Gretchen Krueger, Tom Henighan, Rewon
  Child, Aditya Ramesh, Daniel Ziegler, Jeffrey Wu, Clemens Winter, Chris
  Hesse, Mark Chen, Eric Sigler, Mateusz Litwin, Scott Gray, Benjamin Chess,
  Jack Clark, Christopher Berner, Sam McCandlish, Alec Radford, Ilya Sutskever,
  and Dario Amodei. 2020.
\newblock \href
  {https://proceedings.neurips.cc/paper_files/paper/2020/file/1457c0d6bfcb4967418bfb8ac142f64a-Paper.pdf}
  {Language models are few-shot learners}.
\newblock In \emph{Advances in Neural Information Processing Systems},
  volume~33, pages 1877--1901. Curran Associates, Inc.

\bibitem[{Celikyilmaz et~al.(2020)Celikyilmaz, Clark, and
  Gao}]{celikyilmaz2020evaluation}
Asli Celikyilmaz, Elizabeth Clark, and Jianfeng Gao. 2020.
\newblock Evaluation of text generation: A survey.
\newblock \emph{arXiv preprint arXiv:2006.14799}.

\bibitem[{Chen et~al.(2017)Chen, Fisch, Weston, and
  Bordes}]{chen-etal-2017-reading}
Danqi Chen, Adam Fisch, Jason Weston, and Antoine Bordes. 2017.
\newblock \href {https://doi.org/10.18653/v1/P17-1171} {Reading {W}ikipedia to
  answer open-domain questions}.
\newblock In \emph{Proceedings of the 55th Annual Meeting of the Association
  for Computational Linguistics (Volume 1: Long Papers)}, pages 1870--1879,
  Vancouver, Canada. Association for Computational Linguistics.

\bibitem[{Drozdov et~al.(2022)Drozdov, Wang, Rahimi, McCallum, Zamani, and
  Iyyer}]{drozdov-etal-2022-cant}
Andrew Drozdov, Shufan Wang, Razieh Rahimi, Andrew McCallum, Hamed Zamani, and
  Mohit Iyyer. 2022.
\newblock \href {https://aclanthology.org/2022.findings-emnlp.218} {You can{'}t
  pick your neighbors, or can you? when and how to rely on retrieval in the
  k{NN}-{LM}}.
\newblock In \emph{Findings of the Association for Computational Linguistics:
  EMNLP 2022}, pages 2997--3007, Abu Dhabi, United Arab Emirates. Association
  for Computational Linguistics.

\bibitem[{Grave et~al.(2017)Grave, Joulin, and Usunier}]{grave2017improving}
Edouard Grave, Armand Joulin, and Nicolas Usunier. 2017.
\newblock \href {https://openreview.net/forum?id=B184E5qee} {Improving neural
  language models with a continuous cache}.
\newblock In \emph{International Conference on Learning Representations}.

\bibitem[{Guu et~al.(2020)Guu, Lee, Tung, Pasupat, and Chang}]{guu2020realm}
Kelvin Guu, Kenton Lee, Zora Tung, Panupong Pasupat, and Ming-Wei Chang. 2020.
\newblock {REALM}: Retrieval-augmented language model pre-training.
\newblock In \emph{International Conference on Machine Learning}.

\bibitem[{Hashimoto et~al.(2019)Hashimoto, Zhang, and
  Liang}]{hashimoto-etal-2019-unifying}
Tatsunori~B. Hashimoto, Hugh Zhang, and Percy Liang. 2019.
\newblock \href {https://doi.org/10.18653/v1/N19-1169} {Unifying human and
  statistical evaluation for natural language generation}.
\newblock In \emph{Proceedings of the 2019 Conference of the North {A}merican
  Chapter of the Association for Computational Linguistics: Human Language
  Technologies, Volume 1 (Long and Short Papers)}, pages 1689--1701,
  Minneapolis, Minnesota. Association for Computational Linguistics.

\bibitem[{He et~al.(2022)He, Zhang, and Roth}]{He2022RethinkingWR}
Hangfeng He, Hongming Zhang, and Dan Roth. 2022.
\newblock Rethinking with retrieval: Faithful large language model inference.
\newblock \emph{ArXiv}, abs/2301.00303.

\bibitem[{He et~al.(2021{\natexlab{a}})He, Neubig, and
  Berg-Kirkpatrick}]{he-etal-2021-efficient}
Junxian He, Graham Neubig, and Taylor Berg-Kirkpatrick. 2021{\natexlab{a}}.
\newblock \href {https://doi.org/10.18653/v1/2021.emnlp-main.461} {Efficient
  nearest neighbor language models}.
\newblock In \emph{Proceedings of the 2021 Conference on Empirical Methods in
  Natural Language Processing}, pages 5703--5714, Online and Punta Cana,
  Dominican Republic. Association for Computational Linguistics.

\bibitem[{He et~al.(2021{\natexlab{b}})He, Zhang, Zhou, and
  Glass}]{he-etal-2021-exposure}
Tianxing He, Jingzhao Zhang, Zhiming Zhou, and James Glass. 2021{\natexlab{b}}.
\newblock \href {https://doi.org/10.18653/v1/2021.emnlp-main.415} {Exposure
  bias versus self-recovery: Are distortions really incremental for
  autoregressive text generation?}
\newblock In \emph{Proceedings of the 2021 Conference on Empirical Methods in
  Natural Language Processing}, pages 5087--5102, Online and Punta Cana,
  Dominican Republic. Association for Computational Linguistics.

\bibitem[{Holtzman et~al.(2020)Holtzman, Buys, Du, Forbes, and
  Choi}]{holtzman2019curious}
Ari Holtzman, Jan Buys, Li~Du, Maxwell Forbes, and Yejin Choi. 2020.
\newblock The curious case of neural text degeneration.
\newblock In \emph{International Conference on Learning Representations}.

\bibitem[{Izacard and Grave(2021)}]{izacard-grave-2021-leveraging}
Gautier Izacard and Edouard Grave. 2021.
\newblock \href {https://doi.org/10.18653/v1/2021.eacl-main.74} {Leveraging
  passage retrieval with generative models for open domain question answering}.
\newblock In \emph{Proceedings of the 16th Conference of the European Chapter
  of the Association for Computational Linguistics: Main Volume}, pages
  874--880, Online. Association for Computational Linguistics.

\bibitem[{Izacard et~al.(2022)Izacard, Lewis, Lomeli, Hosseini, Petroni,
  Schick, Dwivedi-Yu, Joulin, Riedel, and Grave}]{izacard_few-shot_2022}
Gautier Izacard, Patrick Lewis, Maria Lomeli, Lucas Hosseini, Fabio Petroni,
  Timo Schick, Jane Dwivedi-Yu, Armand Joulin, Sebastian Riedel, and Edouard
  Grave. 2022.
\newblock \href {http://arxiv.org/abs/2208.03299} {Few-shot {Learning} with
  {Retrieval} {Augmented} {Language} {Models}}.

\bibitem[{Khandelwal et~al.(2021)Khandelwal, Fan, Jurafsky, Zettlemoyer, and
  Lewis}]{khandelwal2021nearest}
Urvashi Khandelwal, Angela Fan, Dan Jurafsky, Luke Zettlemoyer, and Mike Lewis.
  2021.
\newblock Nearest neighbor machine translation.
\newblock In \emph{International Conference on Learning Representations
  (ICLR)}.

\bibitem[{Khandelwal et~al.(2020)Khandelwal, Levy, Jurafsky, Zettlemoyer, and
  Lewis}]{khandelwal20generalization}
Urvashi Khandelwal, Omer Levy, Dan Jurafsky, Luke Zettlemoyer, and Mike Lewis.
  2020.
\newblock {Generalization through Memorization: Nearest Neighbor Language
  Models}.
\newblock In \emph{International Conference on Learning Representations
  (ICLR)}.

\bibitem[{Krishna et~al.(2023)Krishna, Bransom, Kuehl, Iyyer, Dasigi, Cohan,
  and Lo}]{Krishna2023LongEvalGF}
Kalpesh Krishna, Erin Bransom, Bailey Kuehl, Mohit Iyyer, Pradeep Dasigi, Arman
  Cohan, and Kyle Lo. 2023.
\newblock Longeval: Guidelines for human evaluation of faithfulness in
  long-form summarization.
\newblock In \emph{Conference of the European Chapter of the Association for
  Computational Linguistics}.

\bibitem[{Krishna et~al.(2022)Krishna, Chang, Wieting, and
  Iyyer}]{krishna-etal-2022-rankgen}
Kalpesh Krishna, Yapei Chang, John Wieting, and Mohit Iyyer. 2022.
\newblock \href {https://aclanthology.org/2022.emnlp-main.15} {{R}ank{G}en:
  Improving text generation with large ranking models}.
\newblock In \emph{Proceedings of the 2022 Conference on Empirical Methods in
  Natural Language Processing}, pages 199--232, Abu Dhabi, United Arab
  Emirates. Association for Computational Linguistics.

\bibitem[{Krishna et~al.(2021)Krishna, Roy, and
  Iyyer}]{krishna-etal-2021-hurdles}
Kalpesh Krishna, Aurko Roy, and Mohit Iyyer. 2021.
\newblock \href {https://doi.org/10.18653/v1/2021.naacl-main.393} {Hurdles to
  progress in long-form question answering}.
\newblock In \emph{Proceedings of the 2021 Conference of the North American
  Chapter of the Association for Computational Linguistics: Human Language
  Technologies}, pages 4940--4957, Online. Association for Computational
  Linguistics.

\bibitem[{Lan et~al.(2023)Lan, Cai, Wang, Huang, and Mao}]{lan2023copy}
Tian Lan, Deng Cai, Yan Wang, Heyan Huang, and Xian-Ling Mao. 2023.
\newblock \href {https://openreview.net/forum?id=CROlOA9Nd8C} {Copy is all you
  need}.
\newblock In \emph{The Eleventh International Conference on Learning
  Representations}.

\bibitem[{Lee et~al.(2022)Lee, Ping, Xu, Patwary, Fung, Shoeybi, and
  Catanzaro}]{lee2022factuality}
Nayeon Lee, Wei Ping, Peng Xu, Mostofa Patwary, Pascale Fung, Mohammad Shoeybi,
  and Bryan Catanzaro. 2022.
\newblock \href {https://openreview.net/forum?id=LvyJX20Rll} {Factuality
  enhanced language models for open-ended text generation}.
\newblock In \emph{Advances in Neural Information Processing Systems}.

\bibitem[{Lewis et~al.(2020)Lewis, Perez, Piktus, Petroni, Karpukhin, Goyal,
  K\"{u}ttler, Lewis, Yih, Rockt\"{a}schel, Riedel, and Kiela}]{lewis2020rag}
Patrick Lewis, Ethan Perez, Aleksandra Piktus, Fabio Petroni, Vladimir
  Karpukhin, Naman Goyal, Heinrich K\"{u}ttler, Mike Lewis, Wen-tau Yih, Tim
  Rockt\"{a}schel, Sebastian Riedel, and Douwe Kiela. 2020.
\newblock \href
  {https://proceedings.neurips.cc/paper_files/paper/2020/file/6b493230205f780e1bc26945df7481e5-Paper.pdf}
  {Retrieval-augmented generation for knowledge-intensive nlp tasks}.
\newblock In \emph{Advances in Neural Information Processing Systems},
  volume~33, pages 9459--9474. Curran Associates, Inc.

\bibitem[{Mallen et~al.(2023)Mallen, Asai, Zhong, Das, Hajishirzi, and
  Khashabi}]{mallen2023when}
Alex Mallen, Akari Asai, Victor Zhong, Dajarshi Das, Hannaneh Hajishirzi, and
  Daniel Khashabi. 2023.
\newblock When not to trust language models: Investigating effectiveness and
  limitations of parametric and non-parametric memories.
\newblock In \emph{ACL}.

\bibitem[{McCoy et~al.(2021)McCoy, Smolensky, Linzen, Gao, and
  Celikyilmaz}]{McCoy2021HowMD}
R.~Thomas McCoy, Paul Smolensky, Tal Linzen, Jianfeng Gao, and Asli
  Celikyilmaz. 2021.
\newblock How much do language models copy from their training data? evaluating
  linguistic novelty in text generation using raven.
\newblock \emph{ArXiv}, abs/2111.09509.

\bibitem[{Merity et~al.(2016)Merity, Xiong, Bradbury, and
  Socher}]{merity2016pointer}
Stephen Merity, Caiming Xiong, James Bradbury, and Richard Socher. 2016.
\newblock \href {http://arxiv.org/abs/1609.07843} {Pointer sentinel mixture
  models}.

\bibitem[{Metzler et~al.(2022)Metzler, Diaz, Zamani, Bendersky, and
  Dehghani}]{metzler2022reml}
Don Metzler, Fernando Diaz, Hamed Zamani, Mike Bendersky, and Mostafa Dehghani.
  2022.
\newblock Retrieval enhanced machine learning.
\newblock In \emph{SIGIR 2022: Proceedings of the 45th International ACM SIGIR
  Conference on Research and Development in Information Retrieval (Perspectives
  Track)}.

\bibitem[{Mialon et~al.(2023)Mialon, Dess{\`i}, Lomeli, Nalmpantis, Pasunuru,
  Raileanu, Rozi{\`e}re, Schick, Dwivedi-Yu, Celikyilmaz, Grave, LeCun, and
  Scialom}]{Mialon2023AugmentedLM}
Gr{\'e}goire Mialon, Roberto Dess{\`i}, Maria Lomeli, Christoforos Nalmpantis,
  Ramakanth Pasunuru, Roberta Raileanu, Baptiste Rozi{\`e}re, Timo Schick, Jane
  Dwivedi-Yu, Asli Celikyilmaz, Edouard Grave, Yann LeCun, and Thomas Scialom.
  2023.
\newblock Augmented language models: a survey.
\newblock \emph{ArXiv}, abs/2302.07842.

\bibitem[{Nan et~al.(2021)Nan, Nallapati, Wang, Nogueira~dos Santos, Zhu,
  Zhang, McKeown, and Xiang}]{nan-etal-2021-entity}
Feng Nan, Ramesh Nallapati, Zhiguo Wang, Cicero Nogueira~dos Santos, Henghui
  Zhu, Dejiao Zhang, Kathleen McKeown, and Bing Xiang. 2021.
\newblock \href {https://doi.org/10.18653/v1/2021.eacl-main.235} {Entity-level
  factual consistency of abstractive text summarization}.
\newblock In \emph{Proceedings of the 16th Conference of the European Chapter
  of the Association for Computational Linguistics: Main Volume}, pages
  2727--2733, Online. Association for Computational Linguistics.

\bibitem[{Novikova et~al.(2017)Novikova, Du{\v{s}}ek, Cercas~Curry, and
  Rieser}]{novikova-etal-2017-need}
Jekaterina Novikova, Ond{\v{r}}ej Du{\v{s}}ek, Amanda Cercas~Curry, and Verena
  Rieser. 2017.
\newblock \href {https://doi.org/10.18653/v1/D17-1238} {Why we need new
  evaluation metrics for {NLG}}.
\newblock In \emph{Proceedings of the 2017 Conference on Empirical Methods in
  Natural Language Processing}, pages 2241--2252, Copenhagen, Denmark.
  Association for Computational Linguistics.

\bibitem[{Pillutla et~al.(2021)Pillutla, Swayamdipta, Zellers, Thickstun,
  Welleck, Choi, and Harchaoui}]{Pillutla2021MAUVEMT}
Krishna Pillutla, Swabha Swayamdipta, Rowan Zellers, John Thickstun, Sean
  Welleck, Yejin Choi, and Za{\"i}d Harchaoui. 2021.
\newblock Mauve: Measuring the gap between neural text and human text using
  divergence frontiers.
\newblock In \emph{Neural Information Processing Systems}.

\bibitem[{Pimentel et~al.(2023)Pimentel, Meister, and
  Cotterell}]{pimentel2023on}
Tiago Pimentel, Clara~Isabel Meister, and Ryan Cotterell. 2023.
\newblock \href {https://openreview.net/forum?id=bvpkw7UIRdU} {On the
  usefulness of embeddings, clusters and strings for text generation
  evaluation}.
\newblock In \emph{The Eleventh International Conference on Learning
  Representations}.

\bibitem[{Radford et~al.(2019)Radford, Wu, Child, Luan, Amodei, and
  Sutskever}]{radford2019language}
Alec Radford, Jeff Wu, Rewon Child, David Luan, Dario Amodei, and Ilya
  Sutskever. 2019.
\newblock Language models are unsupervised multitask learners.

\bibitem[{Rae et~al.(2021)Rae, Borgeaud, Cai, Millican, Hoffmann, Song,
  Aslanides, Henderson, Ring, Young, Rutherford, Hennigan, Menick, Cassirer,
  Powell, van~den Driessche, Hendricks, Rauh, Huang, Glaese, Welbl, Dathathri,
  Huang, Uesato, Mellor, Higgins, Creswell, McAleese, Wu, Elsen, Jayakumar,
  Buchatskaya, Budden, Sutherland, Simonyan, Paganini, Sifre, Martens, Li,
  Kuncoro, Nematzadeh, Gribovskaya, Donato, Lazaridou, Mensch, Lespiau,
  Tsimpoukelli, Grigorev, Fritz, Sottiaux, Pajarskas, Pohlen, Gong, Toyama,
  de~Masson~d'Autume, Li, Terzi, Mikulik, Babuschkin, Clark, de~Las~Casas, Guy,
  Jones, Bradbury, Johnson, Hechtman, Weidinger, Gabriel, Isaac, Lockhart,
  Osindero, Rimell, Dyer, Vinyals, Ayoub, Stanway, Bennett, Hassabis,
  Kavukcuoglu, and Irving}]{Rae2021ScalingLM}
Jack~W. Rae, Sebastian Borgeaud, Trevor Cai, Katie Millican, Jordan Hoffmann,
  Francis Song, John Aslanides, Sarah Henderson, Roman Ring, Susannah Young,
  Eliza Rutherford, Tom Hennigan, Jacob Menick, Albin Cassirer, Richard Powell,
  George van~den Driessche, Lisa~Anne Hendricks, Maribeth Rauh, Po-Sen Huang,
  Amelia Glaese, Johannes Welbl, Sumanth Dathathri, Saffron Huang, Jonathan
  Uesato, John F.~J. Mellor, Irina Higgins, Antonia Creswell, Nathan McAleese,
  Amy Wu, Erich Elsen, Siddhant~M. Jayakumar, Elena Buchatskaya, David Budden,
  Esme Sutherland, Karen Simonyan, Michela Paganini, L.~Sifre, Lena Martens,
  Xiang~Lorraine Li, Adhiguna Kuncoro, Aida Nematzadeh, Elena Gribovskaya,
  Domenic Donato, Angeliki Lazaridou, Arthur Mensch, Jean-Baptiste Lespiau,
  Maria Tsimpoukelli, N.~K. Grigorev, Doug Fritz, Thibault Sottiaux, Mantas
  Pajarskas, Tobias Pohlen, Zhitao Gong, Daniel Toyama, Cyprien
  de~Masson~d'Autume, Yujia Li, Tayfun Terzi, Vladimir Mikulik, Igor
  Babuschkin, Aidan Clark, Diego de~Las~Casas, Aurelia Guy, Chris Jones, James
  Bradbury, Matthew~G. Johnson, Blake~A. Hechtman, Laura Weidinger, Iason
  Gabriel, William~S. Isaac, Edward Lockhart, Simon Osindero, Laura Rimell,
  Chris Dyer, Oriol Vinyals, Kareem~W. Ayoub, Jeff Stanway, L.~L. Bennett,
  Demis Hassabis, Koray Kavukcuoglu, and Geoffrey Irving. 2021.
\newblock Scaling language models: Methods, analysis \& insights from training
  gopher.
\newblock \emph{ArXiv}, abs/2112.11446.

\bibitem[{Ram et~al.(2023)Ram, Levine, Dalmedigos, Muhlgay, Shashua,
  Leyton-Brown, and Shoham}]{ram2023ralm}
Ori Ram, Yoav Levine, Itay Dalmedigos, Dor Muhlgay, Amnon Shashua, Kevin
  Leyton-Brown, and Yoav Shoham. 2023.
\newblock \href {https://arxiv.org/abs/2302.00083} {In-context
  retrieval-augmented language models}.

\bibitem[{Shannon(2001)}]{shannon2001mathematical}
Claude~Elwood Shannon. 2001.
\newblock A mathematical theory of communication.
\newblock \emph{ACM SIGMOBILE mobile computing and communications review},
  5(1):3--55.

\bibitem[{Shi et~al.(2023)Shi, Min, Yasunaga, Seo, James, Lewis, Zettlemoyer,
  and tau Yih}]{Shi2023REPLUGRB}
Weijia Shi, Sewon Min, Michihiro Yasunaga, Minjoon Seo, Rich James, Mike Lewis,
  Luke Zettlemoyer, and Wen tau Yih. 2023.
\newblock Replug: Retrieval-augmented black-box language models.
\newblock \emph{ArXiv}, abs/2301.12652.

\bibitem[{Thai et~al.(2022)Thai, Karpinska, Krishna, Ray, Inghilleri, Wieting,
  and Iyyer}]{thai-etal-2022-exploring}
Katherine Thai, Marzena Karpinska, Kalpesh Krishna, Bill Ray, Moira Inghilleri,
  John Wieting, and Mohit Iyyer. 2022.
\newblock \href {https://aclanthology.org/2022.emnlp-main.672} {Exploring
  document-level literary machine translation with parallel paragraphs from
  world literature}.
\newblock In \emph{Proceedings of the 2022 Conference on Empirical Methods in
  Natural Language Processing}, pages 9882--9902, Abu Dhabi, United Arab
  Emirates. Association for Computational Linguistics.

\bibitem[{Tkachenko et~al.(2020-2022)Tkachenko, Malyuk, Holmanyuk, and
  Liubimov}]{LabelStudio}
Maxim Tkachenko, Mikhail Malyuk, Andrey Holmanyuk, and Nikolai Liubimov.
  2020-2022.
\newblock \href {https://github.com/heartexlabs/label-studio} {{Label Studio}:
  Data labeling software}.
\newblock Open source software available from
  https://github.com/heartexlabs/label-studio.

\bibitem[{Trivedi et~al.(2022)Trivedi, Balasubramanian, Khot, and
  Sabharwal}]{Trivedi2022InterleavingRW}
H.~Trivedi, Niranjan Balasubramanian, Tushar Khot, and Ashish Sabharwal. 2022.
\newblock Interleaving retrieval with chain-of-thought reasoning for
  knowledge-intensive multi-step questions.
\newblock \emph{ArXiv}, abs/2212.10509.

\bibitem[{Welleck et~al.(2020)Welleck, Kulikov, Roller, Dinan, Cho, and
  Weston}]{unlikelihood-welleck}
Sean Welleck, Ilia Kulikov, Stephen Roller, Emily Dinan, Kyunghyun Cho, and
  Jason Weston. 2020.
\newblock Neural text generation with unlikelihood training.
\newblock In \emph{8th International Conference on Learning Representations,
  {ICLR} 2020, Addis Ababa, Ethiopia, April 26-30, 2020}.

\bibitem[{Wu et~al.(2022)Wu, Rabe, Hutchins, and Szegedy}]{wu2022memorizing}
Yuhuai Wu, Markus~Norman Rabe, DeLesley Hutchins, and Christian Szegedy. 2022.
\newblock \href {https://openreview.net/forum?id=TrjbxzRcnf-} {Memorizing
  transformers}.
\newblock In \emph{International Conference on Learning Representations}.

\bibitem[{Xu et~al.(2023)Xu, Alon, and Neubig}]{Xu2023WhyDN}
Frank~F. Xu, Uri Alon, and Graham Neubig. 2023.
\newblock Why do nearest neighbor language models work?
\newblock \emph{ArXiv}, abs/2301.02828.

\bibitem[{Yogatama et~al.(2021)Yogatama, de~Masson~d'Autume, and
  Kong}]{Yogatama2021AdaptiveSL}
Dani Yogatama, Cyprien de~Masson~d'Autume, and Lingpeng Kong. 2021.
\newblock Adaptive semiparametric language models.
\newblock \emph{Transactions of the Association for Computational Linguistics},
  9:362--373.

\bibitem[{Zhong et~al.(2022)Zhong, Lei, and Chen}]{Zhong2022TrainingLM}
Zexuan Zhong, Tao Lei, and Danqi Chen. 2022.
\newblock Training language models with memory augmentation.
\newblock In \emph{Conference on Empirical Methods in Natural Language
  Processing}.

\end{thebibliography}
\bibliographystyle{acl_natbib}
\appendix
\section{Appendix}\label{appendix:humen_eval}

\begin{table*}[h]
\small
\centering
\resizebox{0.9\textwidth}{!}{%
\begin{tabular}{m{5cm}m{3cm}m{4cm}m{5cm}}  
 \toprule
 \textbf{Context}  & \textbf{Ground-truth} & \textbf{Most Probable Tokens from \hlc[brilliantlavender]{\textit{base-LM}} vs \hlc[babyblue]{\textit{$k$NN-LM}}} & \textbf{Analysis} \\

  \midrule
  The lyrics were inspired by a story ...... To me , that 's the way a great rock ' n ' roll concert should be : a place where everyone comes together ... Maybe that 's the dream of all art : to break down the barriers and the divisions between
  
  & 
  \textbf{``people"}
  \newline
  \textbf{\hlc[brilliantlavender]{\textit{base-LM}}} probability: 0.26 
  \newline
  \textbf{\hlc[babyblue]{\textit{$k$NN-LM}}} probability: 0.23

  & 
  \textbf{\hlc[brilliantlavender]{\textit{base-LM}}}: 
  \newline 
  ``the"(0.20), ``us"(0.09), ``art"(0.03), ``rock"(0.02)
  \newline
  \textbf{\hlc[babyblue]{\textit{$k$NN-LM}}}:
  \newline 
  ``the"(0.23), ``us"(0.07), ``good"(0.02), `art"(0.02)  
  &
  In this example the \textbf{\hlc[brilliantlavender]{\textit{base-LM}}} predicts the ground-truth \textbf{noun} token ``people" with the highest probability of all tokens (0.26). However, after interpolating with the retrieval distribution, the \textbf{\hlc[babyblue]{\textit{$k$NN-LM}}} decreases the probability of the ground-truth token.  
  \\

  \midrule
  Richmond finished the 1984 season 12th in points , with 11 ...... In the Busch Series , he qualified at the pole position in the two races he entered , and won the Charlotte race . Richmond joined Hendrick Motorsports in 1986 , where he teamed up with veteran crew chief Harry Hyde . It took the team until the middle of the season'
  
  & 
  ``to"
  \newline
  \textbf{\hlc[brilliantlavender]{\textit{base-LM}}} probability: 0.78 
  \newline
  \textbf{\hlc[babyblue]{\textit{$k$NN-LM}}} probability: 0.64

  & 
  \textbf{\hlc[brilliantlavender]{\textit{base-LM}}}:
  \newline 
  ``,"(0.07), ``for"(0.03), ``when"(0.02), `that"(0.02) 
  \newline 
  \textbf{\hlc[babyblue]{\textit{$k$NN-LM}}}: 
  \newline 
  ``,"(0.10), ``for"(0.06), ``."(0.04), ``and"(0.02)

  &
  The ground-truth token to be predicted is the \textbf{preposition} \textbf{``the"}, which the \textbf{\hlc[brilliantlavender]{\textit{base-LM}}} correctly predicts with very high probability. However, the \textbf{\hlc[babyblue]{\textit{$k$NN-LM}}} decreases the probability of the groud-truth token significantly compared to the \textbf{\hlc[brilliantlavender]{\textit{base-LM}}}.
  \\
  \bottomrule
\end{tabular}%
}
\caption{Examples where \knnlm\ hurts the inference of next-token (with different part-of-speech such as noun and preposition) by predicting of the gold token with a lower probability compared to the base-LM}
\label{tab:retrieval_examples}
\end{table*}

\begin{table*}[h]
\small
\centering
\resizebox{0.9\textwidth}{!}{%
\begin{tabular}{@{}p{4cm}p{5cm}p{4.5cm}p{3cm}@{}}
\toprule
\textbf{Context} & \textbf{Continuation} & \textbf{Comment} & \textbf{Error} \\\midrule

[$\ldots$] = = = Shootdown = = = A group of ZIPRA guerrillas, armed with a Strela-2 launcher, waited in the bush beneath Flight 825's flightpath, and & \textbf{Continuation A---$k$NN-LM} two soldiers opened fire on the guerrillas. [$\ldots$] The ZIPRA and ZIPRA units exchanged fire, and several ZIPRA members were shot. After the fire was over, the ZIPRA and ZIPRA guerrillas returned fire and began shooting back, and one of the ZIPRA unsuccessfully attacked a ZAPU base at the head of Flight 825, forcing the major to retreat. = = = = ZAPU retreat = = = =   = = = = ZAPU retreat = = = = [$\ldots$] The ZIPRA moved up the hill and captured several ZIPRA members, forcing them to withdraw. [$\ldots$] & Both of these continuations have problems. I chose A because its problems seems to influence the meaning less than the problems in B. [$\ldots$] Continuation A has some repeated words and in a few spots ZIPRA and ZAPU seem to get mixed up, leading to sentences like this: "The ZIPRA moved up the hill and captured several ZIPRA members, forcing them to withdraw." It was easier to recover from those confusions than from the ones in Continuation B. & Repetitive content; contradiction \\\midrule

Hu was born in <unk> County, Anhui Province in 1584 or early 1585. [$\ldots$] By 1619, Hu had moved to Nanjing where he lived with his wife Wu.  & \textbf{Continuation B---GPT-2} During this time he also became a leading scholar of the court, a position he held until 1624, [$\ldots$]. His last year at the academy was 1715. In 1725, Hu was appointed secretary of the government [$\ldots$]. In 1727, he was appointed as vice-president of the Academy, and he remained the academy's secretary until 1742, [$\ldots$].  & I selected that this was a difficult decision [$\ldots$]. However, Continuation B is an impossible choice as it has Hu being born in 1584 and still being alive in 1725 which of course is impossible. I was left with process of elimination in my opinion. & Illogical; wrong chronological details\\\midrule

The lanes of US 75, which are on the bottom level, are 20 feet (6m) below ground level, giving the structure a total height of 140 feet from bottom to top As part of the project, I-635 was widened to include four dedicated high-occupancy vehicle lanes (HOV lanes) that are barrier-separated and reversible. The roads on the five levels are: Level I: US 75, an eight lane highway, four lanes going each way; Level II: The junction of the two six-lane frontage roads, each having three through & \textbf{Continuation A---$k$NN-LM} traffic lanes with one lane in each direction. Level III : In the northbound direction, US 75 curves to the northeast and US 75 curves to the northwest to follow a more southerly routing. The northbound lanes are two lanes in each direction, while the southbound lanes are two lanes. [$\ldots$] = = Major intersections = = [$\ldots$] = Battle of the Kalka River = The Battle of the Kalka River (Polish : Kalka, pronounced [\textprimstress ka\textupsilon], or [\textprimstress ka\textupsilon \textprimstress ma\textupsilon]) was a decisive battle of the 15th and 16th centuries of the 15th century in Austrian occupied Poland. The battle was fought in the middle of the Russo & Continuation A gives more information and is more specific than B. It has some lack of clarity -- Level III has the road curving both northeast and northwest -- but overall it's more informative than just "the intersection of two frontage roads." I would have ruled out A because of the Battle of the Kalka River section at the bottom. It doesn't really have a place in this text. But B also has the Battle of the Falkland Islands, so that's a wash. A it is. & Abrupt topic change; illogical \\\midrule

(Same as above) & \textbf{Continuation B---GPT-2} traffic lanes. Level III : The intersection of the two six-lane frontage roads, each having four through traffic lanes. Level IV: The intersection of the two six-lane frontage roads, each having two through traffic lanes. Level V: The intersection of the two six-lane frontage roads, each having two through traffic lanes = = = Construction = = = The project was approved in July 1997 and completed on December 17, 1997, with the groundbreaking ceremony taking place at the State University of New York (SUNY) campus in New York City. [$\ldots$] = = = Major intersections = = = [$\ldots$] = Battle of the Falkland Islands = & (Same as above) & Abrupt topic change; repetitive content \\\bottomrule
\end{tabular}%
}
\caption{Selected representative examples with evaluators' comments. The error categories in the last column are ours. Contents are shortened for the sake of space.}
\label{tab:human_eval_error}
\end{table*}

\begin{figure*}
    \centering
    \includegraphics[scale=0.2]{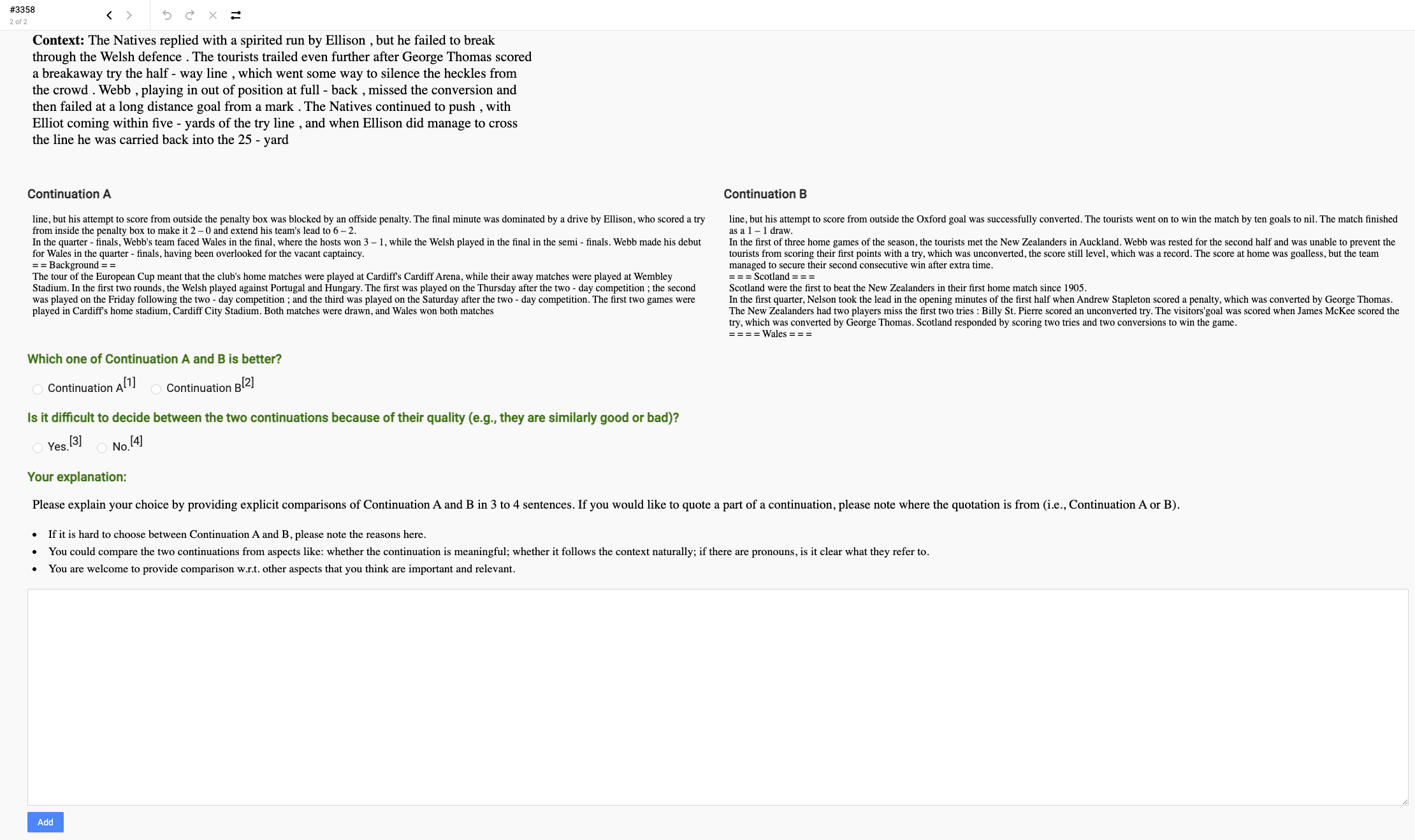}
    \caption{The interface of the human evaluation. Each task consists of a context text, two continuations, two choices, and a free-form justification text box.}
    \label{fig:label_studio_screenshot}
\end{figure*}

\end{document}